\documentclass[sigconf]{acmart}
\usepackage{multirow}
\usepackage{tabularx}
\usepackage{makecell}
\usepackage{float}
\usepackage{pifont} 
\usepackage{hyperref}

\AtBeginDocument{%
  }

\setcopyright{acmlicensed}
\copyrightyear{2025}
\acmYear{2025}
\acmDOI{XXXXXXX.XXXXXXX}
\acmConference[CODS '25]{Make sure to enter the correct
  conference title from your rights confirmation email}{Dec 17--20,
  2025}{Pune, India}
\acmBooktitle{Proceedings of the 12th ACM IKDD Conference on Data Science}
\acmISBN{978-1-4503-XXXX-X/2018/06}

\begin{document}

\title{Process Integrated Computer Vision for Real-Time Failure Prediction in Steel Rolling Mill}

\author{Vaibhav Kurrey}
\email{vaibhavkurrey@iitbhilai.ac.in}
\affiliation{%
  \institution{Indian Institute of Technology}
  \city{Bhilai}
  \state{Chhattisgarh}
  \country{India}
}

\author{Sivakalyan Pujari}
\email{pujaris@iitbhilai.ac.in}
\affiliation{%
  \institution{Indian Institute of Technology}
  \city{Bhilai}
  \state{Chhattisgarh}
  \country{India}
}

\author{Gagan Raj Gupta}
\email{gagan@iitbhilai.ac.in }
\affiliation{%
  \institution{Indian Institute of Technology}
  \city{Bhilai}
  \state{Chhattisgarh}
  \country{India}
}

\renewcommand{\shortauthors}{Vaibhav et al.}

\begin{abstract}
\begin{sloppypar}
We present a long-term deployment study of Machine Vision-based Anomaly Detection for failure prediction in a steel rolling mill. The core of our approach integrates industrial cameras to monitor visual cues such as equipment operation, alignment, and hot bar motion in real time down the process line. These live video streams are processed using deep learning models deployed on a centralized video server system. This computer vision setup allows for prediction of equipment failures and potential process interruptions,  thereby saving huge manufacturing unplanned breakdown costs.  Efficient server-based inference reduces computational load on centralized industrial process control systems (Programmed Logic Controllers) for monitoring, enabling seamless scalability across the production line with minimal resources and reduced risk of computational  loading related process interruptions. The system simultaneously analyzes  sensor data from Data Acquisition systems and visual streams from cameras, identifying both the location and potential root causes of failures. In addition,  it provides actionable insights to operators for  proactive maintenance planning, significantly improving reliability, productivity, and profitability.
\end{sloppypar}
\end{abstract}




\maketitle
\newpage
\section{Introduction}

Industrial setups are complex technical environments where huge machinery, complex control systems, a large number of sensors for feedback and actuators for actions are involved. The process of making an item itself involves multiple small processes. All these processes and sensors interact continuously to attain desired quality and quantity of production.  This level of production establishments requires critical monitoring of process and equipment. If the monitoring is done manually it is very difficult to observe the deviations and failures, even performed by most experienced people. Monitoring by experienced personnel is costly and missing the observation is costlier. So, nowadays automation is employed even for such monitoring tasks using alarms and alerts generated by writing programs in the PLCs (Programmed Logic Controllers). But involving PLCs \cite{b1} in these auxiliary tasks is costly,  may slow down the process control system, needs explicit programming knowledge and may lack  generalization between processes. Recently Artificial Intelligence has taken this space which can provide more generalization, less induced complexity in the process, and avoid the need for explicit programming.  With limited resources and available datasets from industrial sensor data, Artificial Intelligence and Machine Learning (AI\&ML) techniques are showing promising results compared to traditional methods of monitoring and process optimization \cite{b2}.
\begin{figure}[h]
    \centering
    \includegraphics[width=1\linewidth]{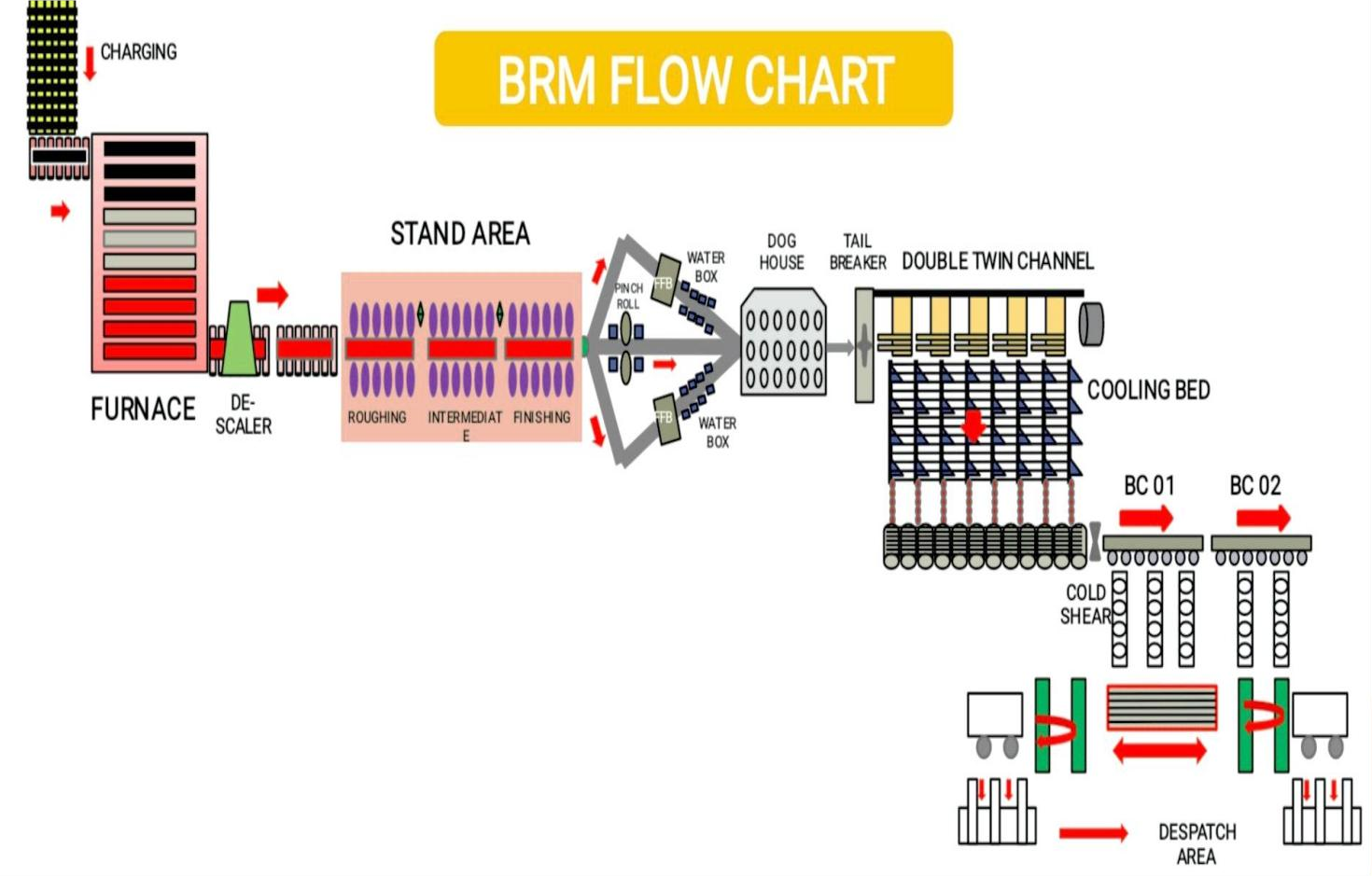} 
    \caption{Steel Bar Rolling Mill Layout}
\end{figure}

\begin{sloppypar}

In the present use case it is a  steel rolling mill, so before diving into our contribution and methodology a brief introduction of the rolling mill in subject also being presented here for readers understanding. The steel Rolling mill chosen is a highly automated mill comprises various group of equipment for rolling a hot square shaped billet of 150 mm x 150 mm and 12 m length is reduced/rolled to round shaped TMT bars of various diameters ranging  from 8 sq.mm to 32 sq.mm and various commercial lengths of 12 m, 9 m bundles. While reducing these billets to desired commercial dimensions, various motors, rollers are employed  for compression, elongation and loopers for releasing the internal stresses developed inside the bar during rolling at stand area as shown in layout at Figure 1 . Further the desired output bars are thermo mechanically treated at water boxes for mechanical properties  and  cut to  dividing lengths which are generally multiples of commercial lengths and diverted to 4 different lines at dog house for discharge on to  the cooling bed  where the bars are naturally cooled to attain final desired mechanical properties. In this course of process the loopers and line diverters are of extreme importance and operate in the order of 100-200 ms, whose malfunctioning causes major stoppages and breakdowns, causing the productivity loss.

\end{sloppypar}

 In present steel market  scenario maintaining uninterrupted operations and high product quality is critical to productivity and profitability. However,  the harsh industrial environment, high-speed mechanical movements, and frequent equipment wear out make failure prediction and timely maintenance is a complex and pressing challenge. Traditional anomaly detection systems in such environments primarily rely on time series data from sensors such as torque, RPM, current, and temperature. Although effective to some extent, these sensor-based systems often suffer from limited context awareness and lack the capability to visually assess physical issues such as equipment malfunctions, misalignments, surface defects, or mechanical deformations that may precede failures. Furthermore, with increasing automation and demand for real-time responsiveness,  it becomes essential to adopt smarter, faster, and more scalable monitoring solutions.

Machine vision, powered by industrial-grade cameras and deep learning algorithms, has emerged as a promising solution to address these limitations. By visually capturing equipment behavior and material flow. Machine vision enables the detection of subtle changes that are difficult to quantify with sensor data alone such as equipment malfunctions, misaligned guide conveyors, foreign/unwanted objects on the process line, or minor cracks that can evolve into critical faults. When coupled with intelligent processing, these systems can recognize patterns and anomalies that are indicative of impending failures \cite{b3}, \cite{b4} . However, deploying such systems in real-time industrial settings comes with its own challenges, particularly related to integration with the existing process control systems, data acquisition systems, alert and monitoring systems without increasing the  computational overhead on the existing systems, avoid data transfer latency while ensuring scalability across multiple production lines \cite{b5}, \cite{b6} .

To overcome these barriers, we propose a practical and scalable machine vision-based anomaly Detection framework that combines time-series data with visual intelligence, designed specifically for deployment as an independent video processing server system. Our system performs real-time inference by seamless integration with the process control systems and data acquisition systems without causing any computational overhead on them. This seamless and simple  integration not only results in robust system architecture but also ensures low-latency decision-making in manufacturing process, crucial for high-speed industrial processes.

The system is capable of continuously analyzing visual streams to detect key cues such as equipment malfunctions, misalignment, and material deviations. It fuses this visual data with conventional sensor streams, allowing it to identify failure patterns more robustly. Beyond detection, it also localizes the region of concern and suggests possible root causes, providing operators with actionable insights that can drive timely and informed maintenance interventions. This hybrid approach, integrating machine vision and sensor data analytics enables a proactive maintenance strategy that improves overall reliability, minimizes unplanned downtime, and enhances operational efficiency.

This paper presents a detailed account of the long-term deployment of our system in a steel rolling mill, outlining the design, implementation, real-world challenges, and observed improvements. Our contributions demonstrate not only the technical feasibility of data integrated Machine Vision in industrial settings but also its tangible impact on failure prediction accuracy, response time, and operational scalability.
\section{Related Work}

The application of anomaly detection in industrial settings has been widely explored, primarily using time series data from sensors embedded in equipment\cite{b7},\cite{b8}.  Traditional methods such as Statistical Process Control (SPC), Principal Component Analysis (PCA), and Autoencoders have demonstrated effectiveness in identifying process deviations \cite{b9}. However, these approaches often fall short in capturing visual anomalies or mechanical misalignments that are not reflected in sensor readings alone.

Recent advances in deep learning have paved the way for machine vision to be adopted in manufacturing and quality control. Convolutional Neural Networks (CNNs) and Vision Transformers (ViTs) have been successfully used for defect detection in steel surfaces \cite{b10}, weld inspection \cite{b11}, real-time tracking of components in high-speed production lines \cite{b12}, visual analytics for analyzing the parameters causing the air entrap in high pressure die casts\cite{b13}, visual analytics in condition monitoring\cite{b14}, sequence tracking and process monitoring\cite{b15}, and for manufacturing process monitoring in smoke detector sensor manufacturing\cite{b16}.

Even though, Within the steel industry, limited research has focused on integrating both sensor data and machine vision for holistic anomaly detection. Most existing systems either rely on control loop signals or post-failure analysis, missing the opportunity for proactive intervention. 

Our work aims to bridge these gaps by combining continuous sensor data streams with real-time machine vision analysis on edge devices, thereby enabling timely, interpretable, and scalable failure prediction for harsh industrial settings.

\section{Methodology}

The proposed framework for process-integrated computer vision in steel rolling mills is designed as an independent, real-time anomaly detection and monitoring system. The methodology is structured into the following major components: (i) system architecture, (ii) camera acquisition and preprocessing, (iii) frame buffering and video storage, (iv) real-time machine vision inference and analytics, (v) sensor data fusion, (vi) alert generation with database integration, (vii) integration with process control systems, (viii) web-based visualization, and (ix) scalability.  
\subsection{System Architecture Overview}
The system integrates industrial-grade cameras, a centralized video server equipped with GPU/CPU resources, and the plant’s existing data acquisition systems. Cameras continuously capture visual streams of hot bar motion, roller alignment, and auxiliary equipment operation. These video feeds are transmitted to the server, where deep learning-based analytics are executed in real time.  

\begin{figure}[h]
    \centering
    \includegraphics[width=0.7\linewidth]{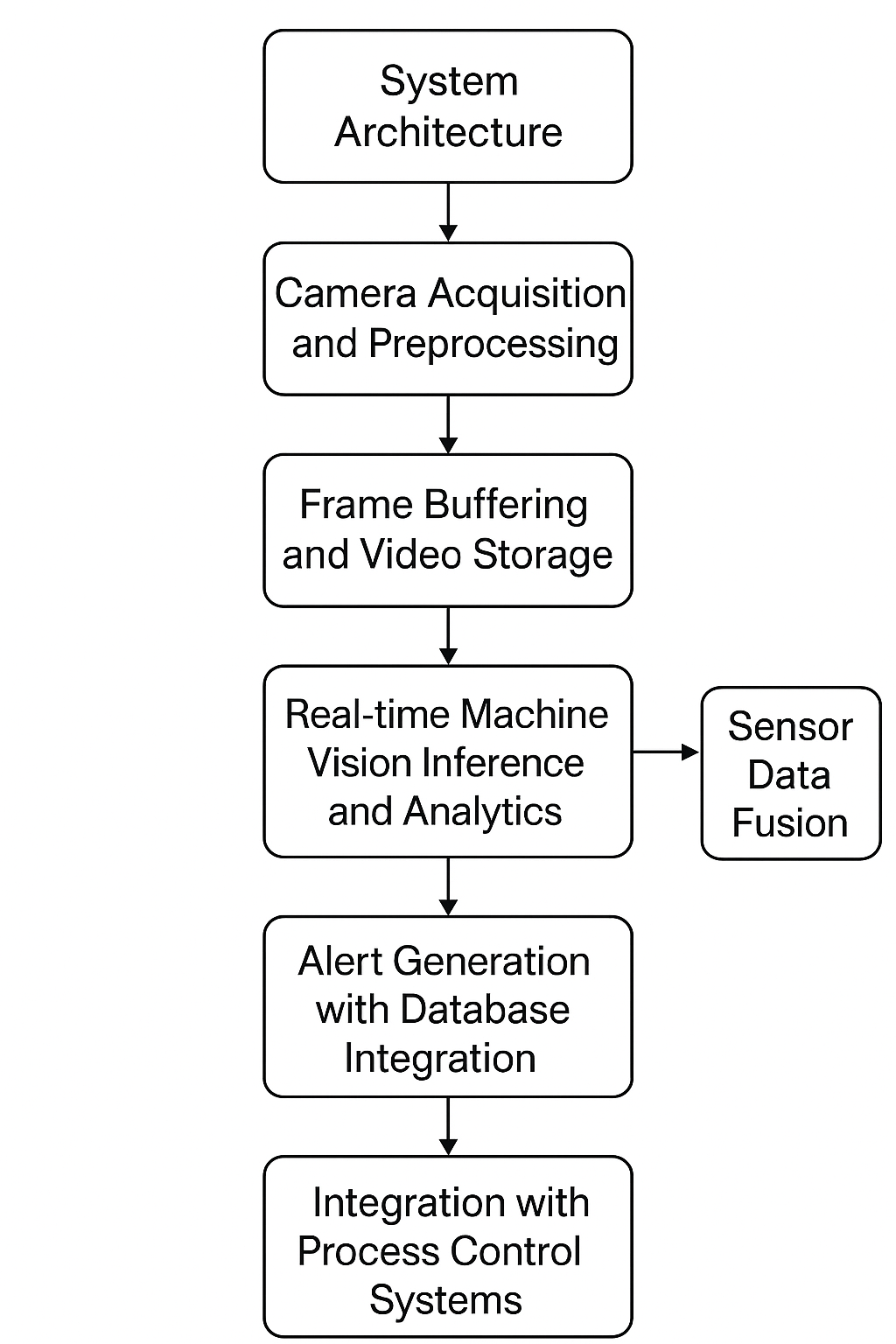} 
    \caption{System architecture and workflow of the proposed process-integrated computer vision framework in steel rolling mills.}
\end{figure}

\subsection{Camera Acquisition and Preprocessing}
High-speed industrial Baumer cameras were strategically installed at critical points in the rolling mill. The cameras were configured using the NeoAPI SDK with controlled exposure times to handle harsh illumination conditions. Video frames were transferred into system memory using custom buffer management, which allowed allocation in CPU memory or GPU memory depending on processing requirements.  

Each acquired frame was validated for pixel format consistency (e.g., BayerRG8) and converted into RGB format. A stable acquisition loop ensured uninterrupted frame streaming at approximately 45 FPS. The framework included redundancy mechanisms for automatic camera reconnection in case of disconnection.  

\subsection{Frame Buffering and Video Storage}
To manage high frame rates, a thread-safe buffer queue was employed to synchronize data transfer between the acquisition and processing modules. For archival purposes, continuous video streams were segmented into two-minute clips, organized hierarchically by date, time, and rod profile size. This ensured both real-time monitoring and offline inspection, without frame loss.  

\subsection{Real-time Machine Vision Inference and Analytics}
The detection module employed YOLO-based deep learning models optimized for GPU execution, to identify and track objects of interest in the mill environment. Separate pre-trained weights were used for different rod profiles (10 mm, 12 mm, 16 mm, 20 mm, and 40 mm), ensuring profile-specific robustness.  

From each frame, multiple features were extracted:  

\begin{itemize}
    \item \textbf{Rod detection and vibration analysis:} The rod’s center coordinates were tracked over time. Vertical displacement patterns were monitored to detect vibrations exceeding predefined thresholds.  
    \item \textbf{Flapper tracking:} Flapper displacement was measured against static baseline coordinates, with deviations logged as anomalies.  
    \item \textbf{Diverter shift measurement:} Diverter positions were detected, and pixel-level shifts were converted into millimeter displacements using a calibration factor.  
    \item \textbf{Rod presence and billet duration:} The system automatically detected rod entry and exit, segmenting billet durations to estimate throughput and identifying short metal events.  
\end{itemize}  

\subsection{Sensor Data Fusion}
Alongside video data, auxiliary process signals from PLC, such as mill operational status, simulation/ghost rolling data, dividing cut data, material presence data at various rolling stages are fetched from IBA based data acqisition system  and integrated to video server through a Redis-based communication layer. This allowed conditional activation of the vision pipeline, where material detections are paused during material absence in the rolling line and intentional manual equipment checks.In addition, dividing cut data used for dynamic suppression of false alerts during different length optimization setups increasing the true positive rate and reliability of the system.

\subsection{Alerting and Database Integration}
All extracted features and statistics were logged into an InfluxDB time-series database. Metrics included camera operating temperature, rod alignment statistics (mean, standard deviation, min, max), flapper displacement, diverter shifts, and billet durations.  

An alerting module was developed to trigger real-time notifications in cases of excessive rod vibration, diverter misalignment, or abnormal billet lengths. This ensured actionable insights for operators and minimized response latency.

\subsection{Web-based Visualization}
For real-time monitoring, a FastAPI-based web server streamed the processed video over a browser-accessible endpoint (\texttt{/video\_feed}). Operators could visually verify detections and alerts, with bounding boxes and markers overlaid on live video streams. Furthermore, Grafana dashboards, backed by InfluxDB, enabled visualization of long-term trends and process statistics.  

\begin{figure}[h]
    \centering
    \includegraphics[width=1\linewidth]{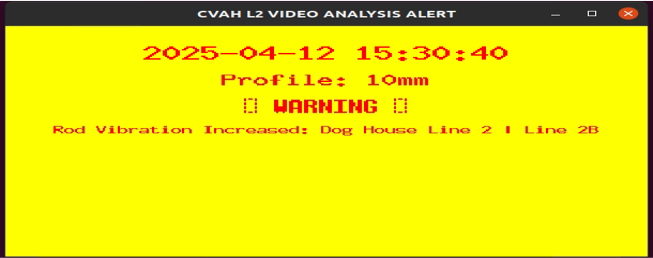} 
    \caption{Real-time alert window displaying vibration and misalignment notification in rolling mill}
\end{figure}

\begin{figure}[h]
    \centering
    \includegraphics[width=1\linewidth]{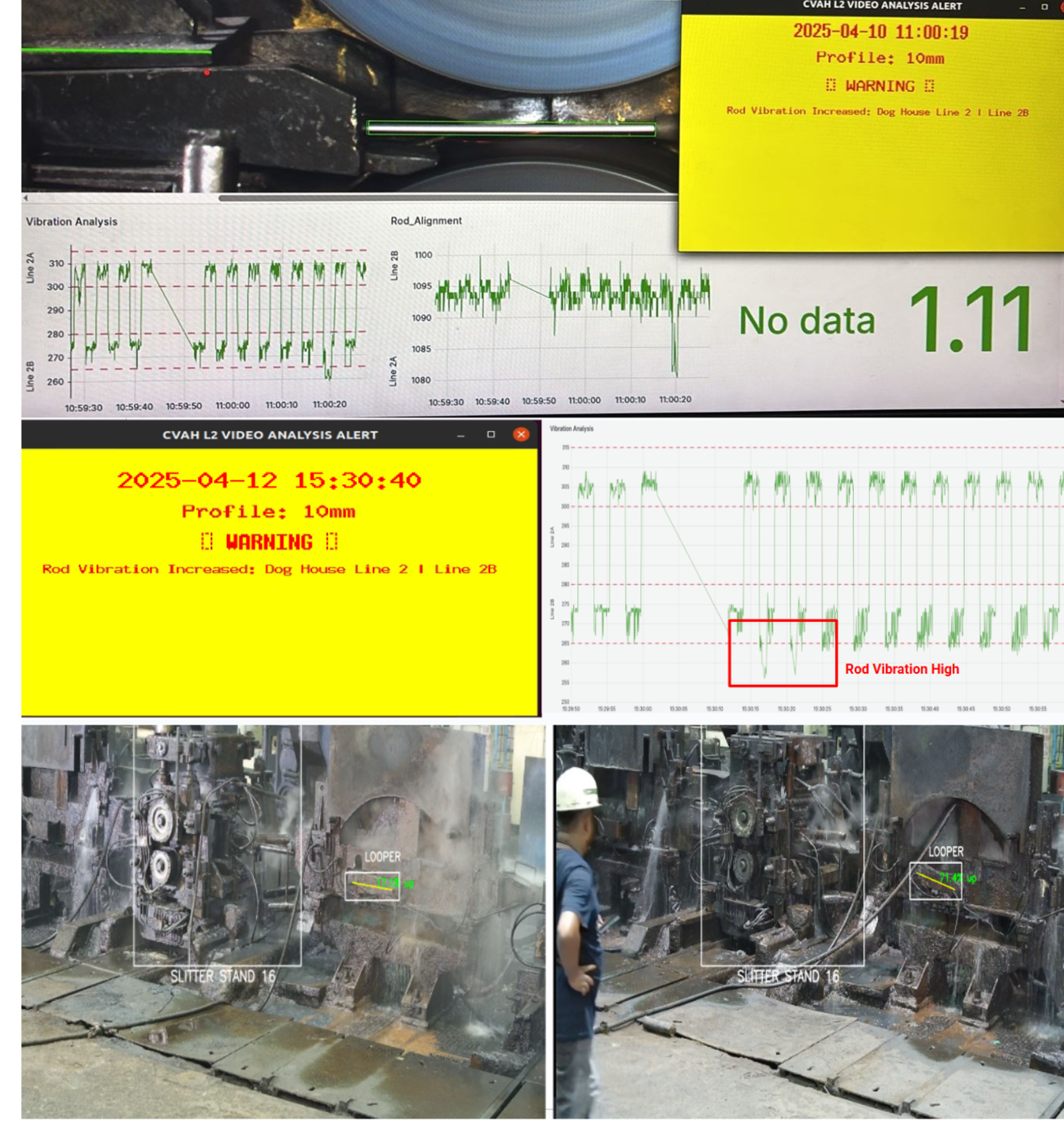} 
    \caption{Web-based integrated visualization Dashboard}
\end{figure}

\subsection{Scalability}
The system was designed with modularity and scalability in mind. Multiple cameras and detection models can be deployed in parallel, with distributed processing supported via GPU acceleration. Storage management ensures long-term data retention while maintaining efficient retrieval for failure analysis.

\section{Results}

The proposed machine vision based anomaly detection framework was deployed in a fully operational steel bar rolling mill for a continuous duration of six months. During this period, multiple operational scenarios—including routine rolling, planned maintenance, and unexpected stoppages were observed to comprehensively evaluate system performance.The data set contains approximately 28,000 labeled frames acquired at 10 FPS across five different rolling dimensions.Ground truth labels were created manually with the help of Domain experts present in the  Rolling Mill. The evaluation focused on four key aspects: (i) detection accuracy, (ii) system latency, (iii) Operational impact, and (iv) scalability. 

\subsection{Detection Accuracy}
The YOLOv11 \cite{b17} based detection models customized for various illumination conditions using 10 fold cross validation  achieved an average precision (mAP@0.5) of 94.2\% across different rolling dimensions.Vibration anomalies were detected with a recall of 92.5\%, while misalignment events (diverter shifts and flapper deviations) recorded a recall of 90.7\%. Rod presence detection achieved near-perfect accuracy ($>$99\%), ensuring reliable billet duration monitoring.  
False positives were reduced by fusion of  visual cues with process signals, resulting in an overall false alarm rate of only 2.8\%.

\subsection{System Latency and Throughput}
Real-time inference was achieved with an average end-to-end latency of 280 ms per frame (including acquisition, pre processing, detection, and alert generation). The system sustained an average processing speed of 42 FPS on the centralized GPU server, meeting the operational requirements of the high-speed rolling process. Integration with Redis and InfluxDB introduced negligible overhead ($<$5 ms), ensuring timely database logging and alerting.

Overall, the results validate the feasibility and effectiveness of deploying process-integrated Machine Vision as an independent server-based monitoring system in harsh steel rolling mill environments. The system not only improves anomaly detection accuracy but also significantly enhances production reliability and operator decision-making. Earlier, the mill faced around 60 cobbles every month, and each cobble caused about 30 minutes of downtime.The system helped prevent nearly 10 cobbles in a month, which directly converts to Rs. 1.15 Cr saving every month at a cost of Rs.21 lakhs per hour of production loss.Some case studies for the readers understanding are also presented at the following drive link
\url{https://drive.google.com/file/d/1_ST8JGHb8tEu7cETlPXyo42s_Q-VRAPb/view?usp=sharing}
\href{https://drive.google.com/file/d/1_ST8JGHb8tEu7cETlPXyo42s_Q-VRAPb/view?usp=sharing}. 

\section{Demonstration}
A video demonstration link is available for review and understanding of reader at:
\url{https://youtu.be/K3A1nbTHH2U?si=gp6txsyR_ry0KCJo}
\href{https://youtu.be/K3A1nbTHH2U?si=gp6txsyR_ry0KCJo}. 
This video showcases the complete process of an integrated computer vision-based solution for pre-failure detection of process equipment. The video showcases: (1) bar rolling process, (2) monitoring of the loopers, doghouse region diverter,  flapper, rod vibration analysis, (3) real-time billet temperature analysis using thermal camera and pyrometer data, (4) visualization of alerts and analytics on the Grafana dashboard.

\newpage

\nobalance

\bibliographystyle{unsrtnat}

\bibliography{references_r2}

\end{document}